\documentclass[conference]{IEEEtran}
\IEEEoverridecommandlockouts
\usepackage{cite}
\usepackage{amsmath,amssymb,amsfonts}
\usepackage{algorithm}
\usepackage{algorithmic}
\usepackage{graphicx}
\usepackage{textcomp}
\usepackage{bm}
\usepackage[table]{xcolor}
\usepackage{booktabs}
\usepackage[caption=false,font=footnotesize]{subfig}
\usepackage{stfloats}

\begin{document}

\title{SCALE-COMM: Shared, Contrastively-Aligned Latent Embeddings for MARL Communication}

\author{
    \IEEEauthorblockN{Mahmoud Abouelyazid, Eman Hammad}
    \\
    \IEEEauthorblockA{\textit{Electrical and Computer Engineering Department} \\
    \textit{Texas A\&M University}, College Station, TX, USA \\
    \{mahmoud75010, eman.hammad\}@tamu.edu}
}

\maketitle

\begin{abstract}
Emergent communication enables partially observant Autonomous Mobile Robots (AMRs) to coordinate effectively in decentralized multi-agent reinforcement learning (MARL) settings. However, existing approaches often struggle with unstable communication protocols, ungrounded message semantics, and interference between communication learning and policy optimization, leading to degraded coordination over time. We propose SCALE-COMM (Shared, Contrastively-Aligned Latent Embeddings for COMMunication), a self-supervised framework for learning compact, stable, and policy-relevant communication representations. SCALE-COMM decouples communication learning from policy optimization by training low-dimensional latent messages that capture task-relevant planning and traffic information, while enforcing consistency across agents and time. Across standard MARL benchmarks and a realistic warehouse coordination task, SCALE-COMM consistently outperforms existing communication frameworks in both representation quality and task performance. The learned communication space yields improved stability, sample efficiency, and throughput under policy fine-tuning, demonstrating the effectiveness of representation-driven communication for scalable multi-agent coordination.
\end{abstract}

\begin{IEEEkeywords}
Self-Supervised Learning, Contrastive Learning, Bandwidth-Efficient Communication, Multi-Agent Reinforcement Learning.
\end{IEEEkeywords}

\section{Introduction}
\label{sec:intro}
Coordinating multiple agents under partial observability is a central MARL challenge. In warehouse and industrial environments, Autonomous Mobile Robots (AMRs) rely on local LiDAR sensing and must cooperate on routing, delivery, and congestion avoidance. Centralized methods exist but scale poorly due to dependence on global map sharing, high bandwidth, and centralized computation, limitations that make them impractical in dynamic, low-bandwidth settings where real-time synchronization is unreliable and local autonomy is required \cite{b1}. This motivates decentralized communication using compact latent messages that encode task intent (routes, goals) and LiDAR-based traffic cues (blocked aisles, congestion zones).

Recent advances include communication-efficient multi-robot collision avoidance \cite{b2}, collaborative LiDAR perception \cite{b3}, and attention-driven communication for dense navigation \cite{b4}. Surveys in autonomous driving MARL \cite{b5} and robust, safe MARL \cite{b6} highlight the need for reliable, bandwidth-efficient, and semantically aligned messages that remain consistently interpretable across agents despite noise and partial observability.

Despite rapid progress, emergent communication methods face three practical obstacles: (i) non-stationarity during joint policy training causes unstable protocols to emerge \cite{b7}; (ii) weak semantic grounding leads to idiosyncratic “dialects” that fail to generalize between agents, time, or environments \cite{b8}, \cite{b9}; and (iii) interference between self-supervised representation learning and control optimization can hinder policy improvement early in training \cite{b10}, \cite{b11}. While recent contrastive methods mitigate some instability, they still cannot preserve message meaning across new tasks or prevent conflicts between representation and control updates \cite{b12}.

We propose SCALE-COMM (Shared, Contrastively-Aligned Latent Embeddings for COMMunication), a self-supervised, curriculum-guided framework for stable and interpretable communication in cooperative MARL. SCALE-COMM aligns message embeddings from multiple agents within a shared latent space through cross-agent, cross-temporal contrastive learning stabilized by exponential-moving-average (EMA) target encoders. A soft curriculum gradually increases self-supervision relative to reinforcement learning, preventing early interference with control optimization. To enhance interpretability, prototype-aligned messaging compresses continuous representations into discrete, token-like embeddings. On the control side, each agent employs a query-conditioned attention mechanism to attend selectively to relevant tasks and traffic.

\begin{figure}[t]
  \centering
  \includegraphics[width=\columnwidth]{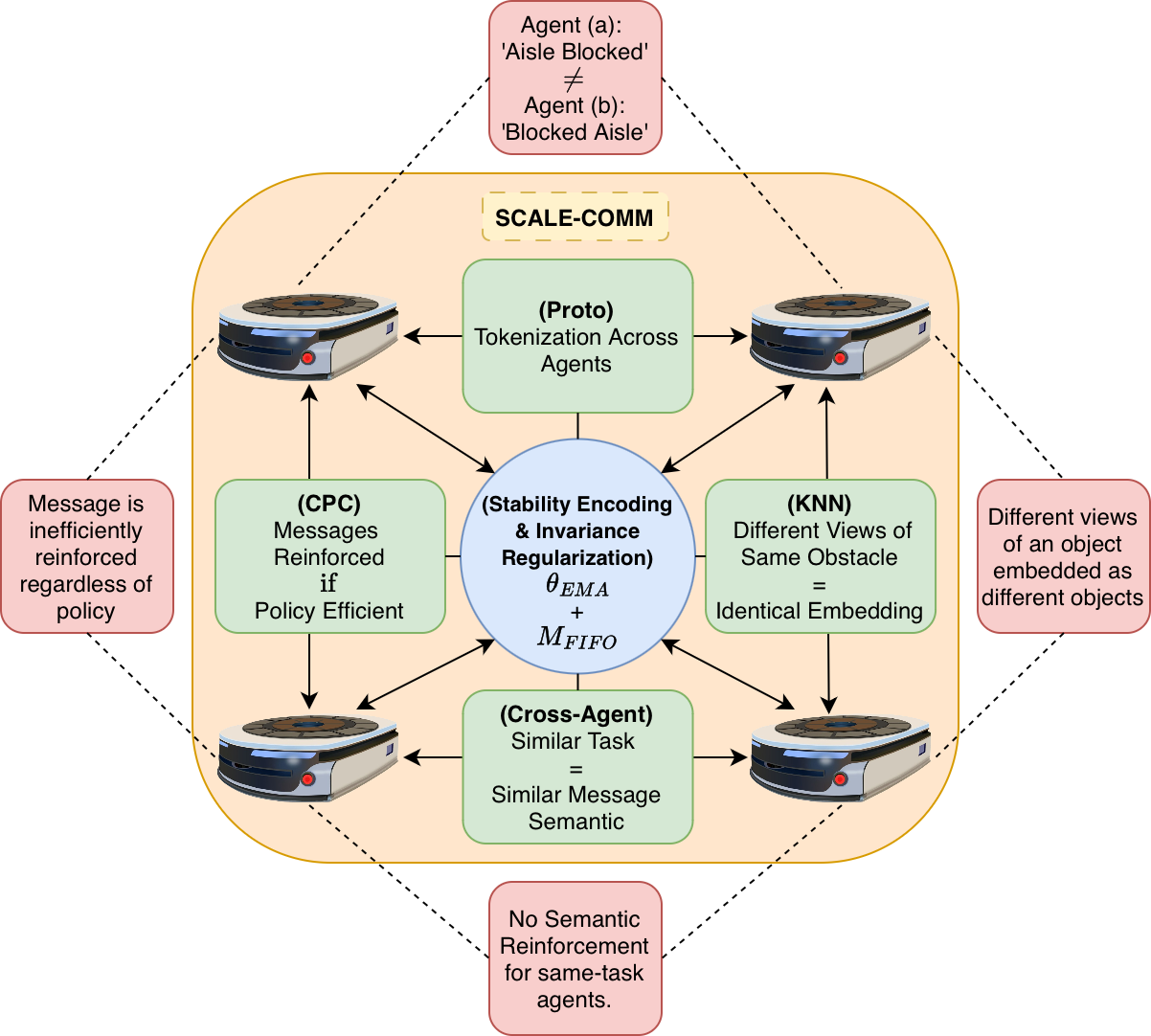}
  \caption{From Caveats to Cures: How SCALE-COMM Fix Message Semantics}
  \label{fig:scale_comm_overview}
\end{figure}

Self-supervision in SCALE-COMM combines complementary objectives that enforce cross-agent alignment, temporal consistency, and semantic abstraction. Contrastive alignment across agents and time grounds messages in shared task context, while temporal prediction and neighborhood consistency reduce representational drift. Prototype-based distillation compresses continuous embeddings into discrete, reusable communication tokens, and invariance regularization improves robustness to noise and viewpoint variation. Training stability is further supported by momentum-based target encoders and a bounded memory of past representations.

The main contri butions of this work are:

\begin{enumerate}
     \item \textbf{SSL-RL Balance:} a unified SSL–RL curriculum that balances representation learning and control optimization;
     
     \item \textbf{Compact and Efficient Representation:} prototype-aligned latent embeddings for compact and interpretable communication.
     
     \item \textbf{Stable Semantic Communication:} a representation-level regularization approach that achieves stable, bandwidth-efficient communication without explicit message pruning.
\end{enumerate}

\section{Related Work}
\label{sec:rel}
Communication learning in cooperative MARL has evolved from early differentiable frameworks toward increasingly selective, contrastive, and self-supervised paradigms. Initial work jointly trained message passing with policy optimization, while subsequent approaches introduced attention and graph-based mechanisms to reduce unnecessary communication. More recent methods leverage self-supervised and contrastive objectives to stabilize representations and align message semantics. The following subsections trace this evolution and outline the remaining challenges in achieving stable, and interpretable communication.

\subsection{Differentiable Communication Frameworks}

Early differentiable communication frameworks such as RIAL, DIAL, and CommNet established that communication protocols can emerge end-to-end within MARL by embedding messages directly into policy optimization \cite{b13,b14}. While effective in small-scale cooperative settings, these methods scale poorly in dynamic or high-dimensional environments. BiCNet extended this idea through a bidirectional recurrent actor–critic architecture \cite{b15}. Subsequent task-agnostic approaches represented multiple observations via differentiable set autoencoders to generalize learned protocols across tasks \cite{b16}. T2MAC introduced selective engagement and evidence-based message filtering, allowing agents to learn both \emph{when} and \emph{to whom} to communicate efficiently \cite{b17}.

\subsection{Targeted Communication}

While differentiable frameworks established the feasibility of emergent communication, they often relied on dense, all-to-all message passing. Selective communication methods addressed this by introducing attention and gating mechanisms that determine \emph{when}, \emph{whom}, and \emph{what} to communicate. ATOC \cite{b18} dynamically forms communication groups via attention; TarMAC \cite{b19} extends this with query–key–value attention for content-specific message exchange. IC3Net \cite{b20} introduces per-agent communication gates for mixed cooperative–competitive tasks, while I2C \cite{b21} leverages probabilistic priors to infer communication relevance. Transformer-based extensions such as TGCNet \cite{b22} and CommFormer \cite{b23} further model inter-agent interactions as dynamic graphs with temporal gating. Together, these works mark a transition from dense, fixed communication to adaptive and context-aware message exchange.

\subsection{Contrastive and Self-Supervised Communication}

Recent research has turned toward discrete and self-supervised protocols to improve interpretability and semantic alignment. Discrete communication methods quantize continuous embeddings into symbolic tokens, promoting compositionality and reducing ambiguity \cite{b24,b25,b26}. In parallel, contrastive self-supervision has been used to stabilize emergent communication by maximizing mutual information between sent and received messages \cite{b27}. For example, Contrastive Alignment for Communication Learning (CACL) treats messages as multiple “views” of an agent’s latent state \cite{b12}, while MASIA applies self-supervised aggregation of received messages into permutation-invariant embeddings \cite{b28}. MA2CL introduces masked attentive contrastive learning for efficient message reconstruction \cite{b29}. Despite these advances, current discrete and contrastive methods remain unstable under non-stationary MARL dynamics, and lack mechanisms for reusable, semantically consistent message embeddings. Those limitations are directly addressed by SCALE-COMM through cross-agent contrastive alignment, prototype-based regularization, and curriculum-driven stabilization.

\begin{figure}[t]
  \centering
  \includegraphics[width=0.9\columnwidth]{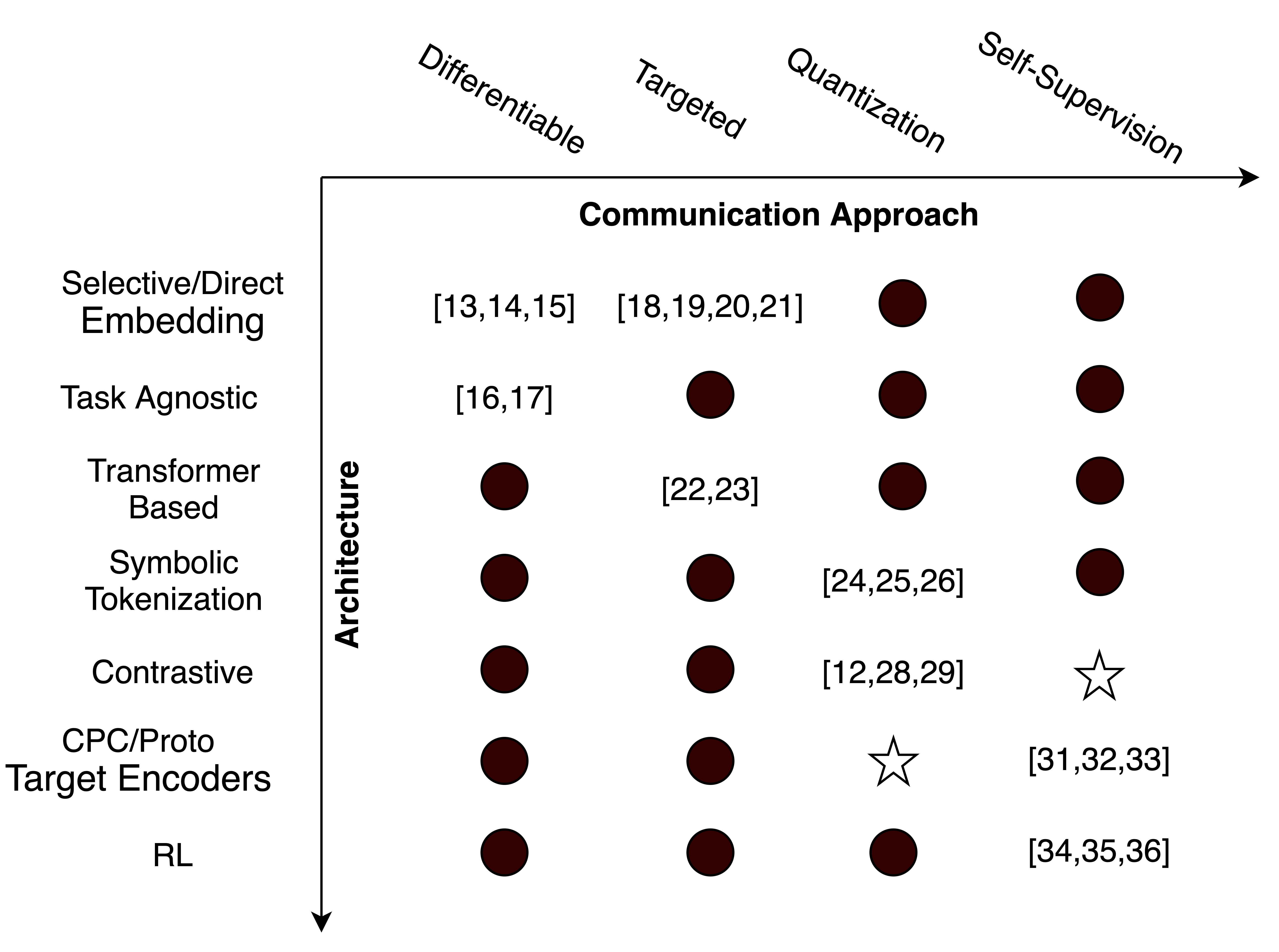}
  \caption{Taxonomy of prior work on communication and self-supervision. This work's contribution is signified by the stars.}
  \label{fig:related_work_overview}
\end{figure}

\subsection{Self-Supervision in Reinforcement Learning}

Advances in self-supervised representation learning have profoundly influenced modern MARL communication. Contrastive Predictive Coding (CPC) introduced temporal InfoNCE objectives for predicting future latents \cite{b30}. MoCo stabilized contrastive learning via a momentum-updated target encoder and a memory queue of negatives \cite{b31}. BYOL eliminated explicit negatives by coupling an online and EMA target encoder \cite{b32}, while SwAV \cite{b33} used prototype-based alignment to form discrete clusters. In reinforcement learning, these ideas inspired auxiliary objectives such as CURL \cite{b34} and SPR \cite{b35}, improving sample efficiency and generalization. AuxDistill further combined auxiliary policy distillation with primary RL objectives to improve long-horizon control \cite{b36}.

\section{Methodology}
\label{sec:metho}
SCALE-COMM unifies key ideas from differentiable communication, bottlenecked architectures, and self-supervised representation learning. As in DIAL and CommNet \cite{b13,b14}, it maintains end-to-end differentiability, but replaces dense message passing with selective, attention-based consumption to control bandwidth and semantic relevance. Like bottlenecked models \cite{b42}, it enforces efficiency, yet achieves this through contrastive and prototype-based objectives rather than structural pruning or fixed compression, yielding compact and discriminative latent channels.

The interpretability of prototype-aligned messages in SCALE-COMM follows from principles established in prototype-based self-supervised learning. Prior work on clustering-based SSL (e.g., SwAV-style methods) shows that introducing a finite set of shared prototypes encourages representations to organize around recurring semantic factors rather than instance-specific noise \cite{b33}. By constraining embeddings to align with learned centroids, prototype objectives reduce the effective degrees of freedom of the representation space and promote the reuse of consistent latent concepts. In multi-agent communication, this inductive bias encourages messages arising from similar coordination contexts to map to the same prototype. Each prototype therefore aggregates message instances across time, agents, and episodes, averaging out stochastic variation while preserving task-relevant structure, yielding representations that are stable, reusable, and semantically meaningful without explicit supervision.

Positioned between self-supervised communication and policy-conditioned representation learning as shown in Fig.~\ref{fig:related_work_overview}, SCALE-COMM incorporates stabilization techniques from BYOL, MoCo, and SwAV \cite{b31,b32,b33}, EMA targets, memory queues, and prototype distillation, to remain stable under non-stationary multi-agent dynamics. It further extends emergent communication and coordination work \cite{b28,b29} by introducing a soft scheduling mechanism that balances contrastive self-supervision with policy optimization, preventing early representational collapse and ensuring that learned protocols remain grounded in task utility.

\begin{figure*}[t]
  \centering
  \includegraphics[width=\textwidth]{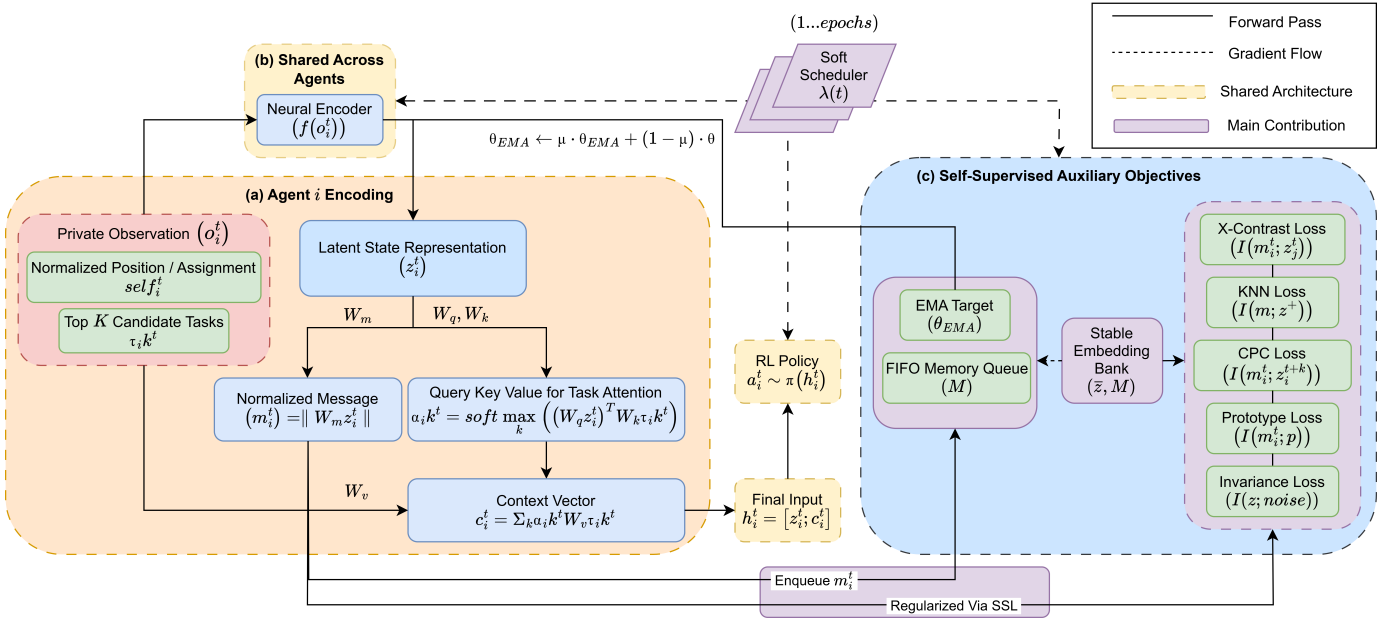}
  \caption{SCALE-COMM architecture. The message space is regularized via self-supervised losses, providing an implicit representation-level bottleneck that stabilizes and compresses communication without explicit pruning}
  \label{fig:scale_comm_architecture}
\end{figure*}

\subsection{Problem Formulation}

We consider a partially observable Markov game (POMG) with $N$ cooperative agents, indexed by $i \in \{1,\dots,N\}$. At each timestep $t$, agent $i$ receives a private observation $o_i^t \in \mathbb{R}^{d}$ composed of (i) self features (normalized position and a carrying flag) and (ii) a fixed-size set of $K$ candidate task descriptors (pickup/drop coordinates and heuristic distances). These features are mapped into a latent embedding:
\begin{equation}
    z_i^t = f_\theta(o_i^t),
\end{equation}
where $f_\theta$ is a shared encoder trained with self-supervised objectives. The latent $z_i^t$ is then projected into a common communication space:
\begin{equation}
    \tilde{m}_i^t = W_m z_i^t, \quad W_m \in \mathbb{R}^{d_m \times d_z},
\end{equation}
and normalized to unit length:
\begin{equation}
    m_i^t = \frac{\tilde{m}_i^t}{\| \tilde{m}_i^t \|_2}.
\end{equation}
This normalization constrains all latent messages to lie on a shared hypersphere, facilitating stable cross-agent comparison and contrastive alignment.

Agent $i$ selects an action $a_i^t \in \{1,\dots,K,K{+}1\}$, where $1{:}K$ index the current top-$K$ task candidates and $K{+}1$ is a skip/no-assignment. Let $\mathbf{a}^t=(a_1^t,\dots,a_N^t)$ denote the joint action. The environment transitions according to $p(s^{t+1}\!\mid s^t,\mathbf{a}^t)$ and provides per-agent shaped rewards $r_i^t$ (e.g., assignment, pickup/drop completion, unassigned penalties). We evaluate cooperative performance via episode-level delivery and utilization KPI and, for learning, optimize the discounted sum of per-agent returns:
\begin{equation}
    J(\theta) \;=\; \mathbb{E}_{\tau \sim \pi_\theta}\!\left[\,\sum_{i=1}^{N}\sum_{t=0}^{T} \gamma^{t}\, r_i^t \right],
\end{equation}
where $\gamma \in [0,1)$ is the discount factor, $\theta$ are the policy parameters, and $\tau$ is the joint trajectory induced by $\{\pi_i\}_{i=1}^N$ and the environment.
\subsection{Attention-Based Message Aggregation}

Although SCALE-COMM does not perform explicit runtime message passing between agents, it implements an implicit attention mechanism that determines which task features and latent cues are most relevant for decision making. After encoding each agent's local observation into a latent representation $z_i^t = f_\theta(o_i^t)$, the policy incorporates a task-level attention bias that selectively weights candidate task descriptors.

Formally, each candidate task feature tensor $\tau_i^t \in \mathbb{R}^{K \times d_\tau}$ is scored by a learned query–key interaction:
\begin{equation}
    \alpha_{ik}^t = \text{softmax}_k\!\big((W_q z_i^t)^\top W_k \tau_{ik}^t\big),
\end{equation}
where $W_q, W_k$ are learnable projections shared across agents. The resulting attention weights $\alpha_{ik}^t$ emphasize the most contextually relevant tasks (e.g., nearby deliveries or high-priority jobs). The attended context vector is then obtained as
\begin{equation}
    c_i^t = \sum_{k=1}^{K} \alpha_{ik}^t W_v \tau_{ik}^t,
\end{equation}
which is concatenated with the agent's latent embedding to form the policy input:
\begin{equation}
    h_i^t = [\,z_i^t ; c_i^t\,].
\end{equation}
During training, SCALE-COMM’s contrastive and prototype-based self-supervision further aligns these attention-driven embeddings across agents, promoting semantic consistency and cooperative coordination.

\subsection{Self-Supervised Auxiliary Objectives}

As shown in Fig.~\ref{fig:scale_comm_architecture}, SCALE-COMM integrates multiple self-supervised losses that jointly shape the geometry of the communication space. These objectives encourage messages to capture task-relevant information, maintain temporal coherence, and remain robust under observation noise.

\paragraph{Cross-Agent Contrast}
Following the CACL approach \cite{b12}, each agent aligns its message embedding with peer latents from similar contexts:
\begin{equation}
    \mathcal{L}_{\text{X-Contrast}} 
    = - \log \frac{\exp(\text{sim}(m_i^t, z_j^t)/\tau)}
    {\sum_{k} \exp(\text{sim}(m_i^t, z_k^t)/\tau)}.
\end{equation}
This InfoNCE loss maximizes $I(m_i^t; z_j^t)$, collapsing messages from agents who perceive similar spatial configurations (e.g., the same blocked aisle) into nearby latent regions.

\paragraph{Nearest-Neighbor Contrast}
To stabilize representation geometry, each message is compared to an exponential moving average (EMA) target and a queue of past embeddings:
\begin{equation}
    \mathcal{L}_{\text{KNN}} 
    = \mathbb{E}_{(m,z^+)} \Big[ - \log 
    \frac{\exp(\text{sim}(m,z^+)/\tau)}
    {\sum_{z^- \in \mathcal{M}} \exp(\text{sim}(m,z^-)/\tau)} \Big].
\end{equation}
By enforcing smooth local consistency, this loss penalizes large embedding shifts for minor sensory variations to yield Lipschitz-continuous representations that remain stable under partial observability and noise.

\paragraph{Temporal CPC}
Predictive consistency is enforced by linking present messages to future latent states within a temporal window $k$:
\begin{equation}
    \mathcal{L}_{\text{CPC}} 
    = - \log \frac{\exp(\text{sim}(g(m_i^t), z_i^{t+k})/\tau)}
    {\sum_{z^- \in \mathcal{M}} \exp(\text{sim}(g(m_i^t), z^-)/\tau)}.
\end{equation}
This contrastive predictive coding term maximizes $I(m_i^t; z_i^{t+k})$, ensuring that current embeddings are temporally predictive to allow agents to implicitly learn to encode short-horizon environmental dynamics such as emerging congestion.

\paragraph{Prototype Distillation.}
Messages are softly aligned to $P$ learnable prototype vectors:
\begin{equation}
    \mathcal{L}_{\text{Proto}} 
    = - \sum_{p=1}^P q_p(m_i^t) 
    \log \frac{\exp(\text{sim}(m_i^t,p)/\tau)}
    {\sum_{j=1}^P \exp(\text{sim}(m_i^t,p_j)/\tau)}.
\end{equation}
This introduces a quantization prior, compressing continuous representations into discrete communication tokens. Conceptually, this acts as an information bottleneck, reducing redundancy and promoting interpretability to ensure distinct warehouse events (e.g., “aisle blocked,” “package picked,” “delivery complete”) map to stable, reusable message prototypes.

\paragraph{Invariance Regularization.}
To ensure robustness under stochastic augmentations and encoder drift, SCALE-COMM aggregates several consistency terms into
\begin{equation}
\mathcal{L}_{\text{inv}} =
    \lambda_1 \, \mathcal{L}_{\text{pred}} +
    \lambda_2 \, \mathcal{L}_{\text{ts}} +
    \lambda_3 \, \mathcal{L}_{\text{hz}} +
    \lambda_4 \, \mathcal{L}_{\text{CKA}},
\end{equation}
where $\mathcal{L}_{\text{pred}}$ aligns online and EMA targets, $\mathcal{L}_{\text{ts}}$ enforces temporal smoothness between consecutive frames, $\mathcal{L}_{\text{hz}}$ encourages cosine similarity between hidden and latent features, and $\mathcal{L}_{\text{CKA}}$ aligns representation geometry via centered kernel alignment. 
\subsection{Stabilization Mechanisms}

To account for non-stationarity during joint optimization of communication and control, SCALE-COMM adopts stabilization techniques from self-supervised learning that regularize encoder dynamics and prevent representation collapse.

\paragraph{Exponential Moving Average (EMA) Targets.}
A momentum-updated target encoder smooths rapid policy-induced shifts:
\begin{equation}
    \theta_{\text{EMA}} \leftarrow \mu \theta_{\text{EMA}} + (1-\mu)\theta,
\end{equation}
where $\mu \in [0,1)$ controls the update rate. The slowly evolving target decouples gradient noise from the online encoder $\theta$, stabilizing positives and ensuring that the InfoNCE denominator changes gradually, as in MoCo and BYOL \cite{b31,b32}.

\paragraph{FIFO Memory Queue and Asymmetric Augmentation.}
A first-in–first-out queue $\mathcal{M}$ stores recent embeddings used as negatives:
\[
\frac{\exp(\text{sim}(m, z^+)/\tau)}{\sum_{z^- \in \mathcal{M}} \exp(\text{sim}(m, z^-)/\tau)},
\]
approximating sampling from the true data distribution and tightening the InfoNCE lower bound on $I(x;y)$. This yields contrastive gradients robust to policy drift. Asymmetric augmentations, such as masking environment observations, perturbing task encodings, and stochastic dropout, further enforce invariance by penalizing deviation between augmented views $\text{sim}(m_i^t, \hat{m}_i^t)$.
\subsection{Final Joint Objective}

SCALE-COMM optimizes a joint objective combining cooperative control with self-supervised regularization to stabilize and structure inter-agent communication.

\paragraph{Reinforcement Learning.}
The control policy is trained via an actor–critic loss:
\begin{equation}
\mathcal{L}_{\text{RL}} =
\mathbb{E}_{\tau \sim \pi_\theta} 
[- \log \pi_\theta(a_i^t \mid h_i^t) A_i^t],
\end{equation}
maximizing expected return by favoring high-advantage actions $A_i^t$.

\paragraph{Self-Supervised Regularization.}
Auxiliary objectives shape the embedding space through alignment, temporal consistency, and discretization:
\begin{equation}
    \mathcal{L}_{\text{SSL}} =
    \alpha \mathcal{L}_{\text{X-Contrast}} +
    \beta \mathcal{L}_{\text{KNN}} +
    \gamma \mathcal{L}_{\text{CPC}} +
    \delta \mathcal{L}_{\text{Proto}} +
    \eta \mathcal{L}_{\text{inv}}.
\end{equation}

Each weight $(\alpha,\beta,\gamma,\delta,\eta)$ was tuned via coarse grid search to balance gradient magnitudes and avoid dominance of any term. We set $(\alpha,\beta,\gamma,\delta,\eta)=(1.0,,0.5,,0.8,,0.3,,0.2)$, emphasizing cross-agent alignment and temporal prediction, while keeping prototype abstraction and invariance as light stabilizers for smooth convergence across environments.

\paragraph{Curriculum Scheduling.}
The total objective is
\begin{equation}
    \mathcal{L} = \mathcal{L}_{\text{RL}} + \lambda(t)\mathcal{L}_{\text{SSL}},
\end{equation}
where $\lambda(t)$ gradually increases during training. This balances early policy learning with later representational regularization, ensuring communication embeddings remain stable and semantically consistent as control improves.

\begin{algorithm}
\caption{SCALE-COMM Training: SSL Pretraining + PPO Fine-tuning}
\label{alg:scalecomm}
\begin{algorithmic}[1]
\fontsize{9pt}{11pt}\selectfont
\REQUIRE Encoder $f_\theta$, EMA target $f_{\theta_{\text{EMA}}}$, queue $\mathcal{M}$, predictor $g$, PPO policy $\pi_\theta$ with value head $V$, task-bias scorer $B$
\STATE \textbf{Setup:} collect replay buffer $\mathcal{D}$ with heuristic policy; set $\theta_{\text{EMA}}\!\leftarrow\!\theta$.

\STATE \textit{Phase I: Self-Supervised Pretraining}
\FOR{each SSL epoch}
  \STATE Sample minibatch $\mathcal{B}\subset\mathcal{D}$; encode $z\!=\!f_\theta(x)$, $z^\star\!=\!f_{\theta_{\text{EMA}}}(x)$ (no grad).
  \STATE Update EMA: $\theta_{\text{EMA}}\!\leftarrow\!\mu\theta_{\text{EMA}}+(1-\mu)\theta$.
  \STATE Compute $\mathcal{L}_{\text{SSL}}
  = \mathcal{L}_{\text{X}} + \mathcal{L}_{\text{KNN}} + \mathcal{L}_{\text{CPC}} + \mathcal{L}_{\text{Proto}} + \mathcal{L}_{\text{inv}}$.
  \STATE Optimize $f_\theta$, $g$, and prototypes; enqueue $z$ in $\mathcal{M}$.
\ENDFOR

\STATE \textit{Phase II: PPO Fine-tuning}
\STATE Initialize policy $\pi_\theta$ with pretrained encoder (freeze or finetune).
\FOR{each PPO iteration}
  \STATE Collect trajectories $(o_i^t,a,r)$ using $\pi_\theta$.
  \IF{$B$ enabled}
    \STATE Apply task bias: $\text{logits}_{1:K}\!+=\!\beta\,B(z,\text{tasks})$.
  \ENDIF
  \STATE Compute PPO objective $\mathcal{L}_{\text{PPO}}$; add temporal SSL term $\mathcal{L}_{\text{aux}}$.
  \STATE Update parameters by minimizing $\mathcal{L}_{\text{PPO}}+\lambda_{\text{aux}}\mathcal{L}_{\text{aux}}$.
\ENDFOR

\STATE \textbf{Return:} Optimized $\pi^{*} = 
\arg\min_\pi \big( \mathcal{L}_{\text{RL}} + 
\lambda(t)\mathcal{L}_{\text{SSL}} \big)$
\STATE (During PPO fine-tuning, $\mathcal{L}_{\text{RL}}\!\approx\!\mathcal{L}_{\text{PPO}}$ and $\mathcal{L}_{\text{SSL}}\!\rightarrow\!\mathcal{L}_{\text{aux}}$)

\end{algorithmic}
\end{algorithm}

From a computational complexity standpoint, during self-supervised pretraining, the dominant cost arises from the cross-agent contrast and temporal consistency modules. This yields a time complexity of \(O(B^{2} d)\) and a memory cost of \(O(B^{2})\), where \(B\) is the batch size and \(d\) is the embedding dimension. Additional linear terms are contributed by the FIFO queue and prototype updates, each scaling as \(O(BQd + BPd)\) for queue size \(Q\) and prototype count \(P\). In the PPO fine-tuning stage, encoder forward passes and policy updates incur \(O(EM d^{2})\) cost per iteration, where \(E\) is the number of epochs and \(M\) the minibatch size, while environment rollouts scale linearly with the number of agents and candidate tasks at \(O(A K d)\) per step. In practice, these quadratic SSL components lead SCALE-COMM to use approximately 77\% more runtime on average than lighter contrastive baselines.

\section{Experiments}
To evaluate SCALE-COMM comprehensively, we design three tiers of experiments.
(1) We benchmark against established baselines (AEComm, CACL, and their differentiable variants AEComm-DIAL and CACL-DIAL \cite{b12,b38}) across standard cooperative control environments: Traffic-Junction, Predator-Prey, and Find-Goal, which test collision avoidance, pursuit–evasion, and goal-directed coordination under partial observability.
(2) We compare against broader self-supervised paradigms, including SwAV \cite{b33}, SimCLR \cite{b39}, and MoCo \cite{b31}, to evaluate embedding stability, alignment quality, and interpretability.
(3) Finally, we evaluate scalability in a custom multi-robot warehouse environment, where agents perform coordinated pick-and-drop operations under dynamic resource constraints, testing protocol adaptability and throughput optimization in realistic industrial settings against the same self-supervised paradigms.

\subsection{Benchmark Environments} 
\begin{figure*}[!t]
  \centering
  \subfloat[Traffic-Junction]{%
    \includegraphics[width=0.31\textwidth]{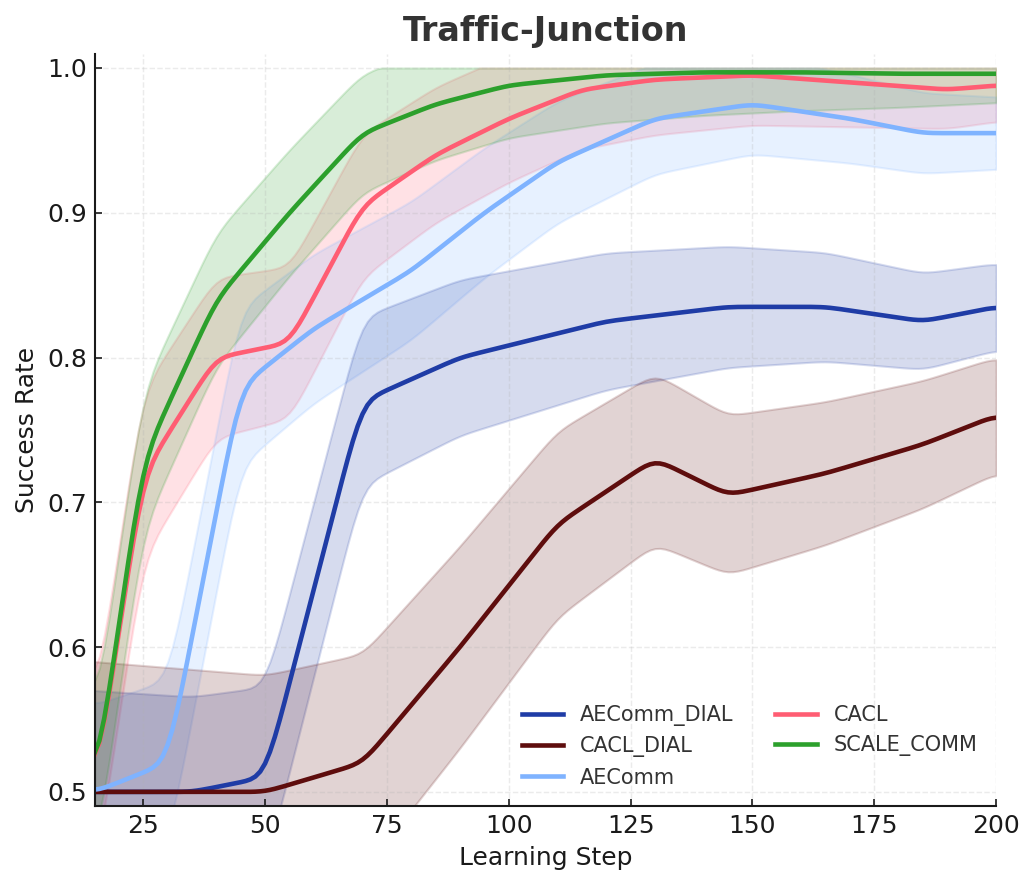}%
    \label{fig:traffic}%
  }\hfill
  \subfloat[Predator-Prey]{%
    \includegraphics[width=0.31\textwidth]{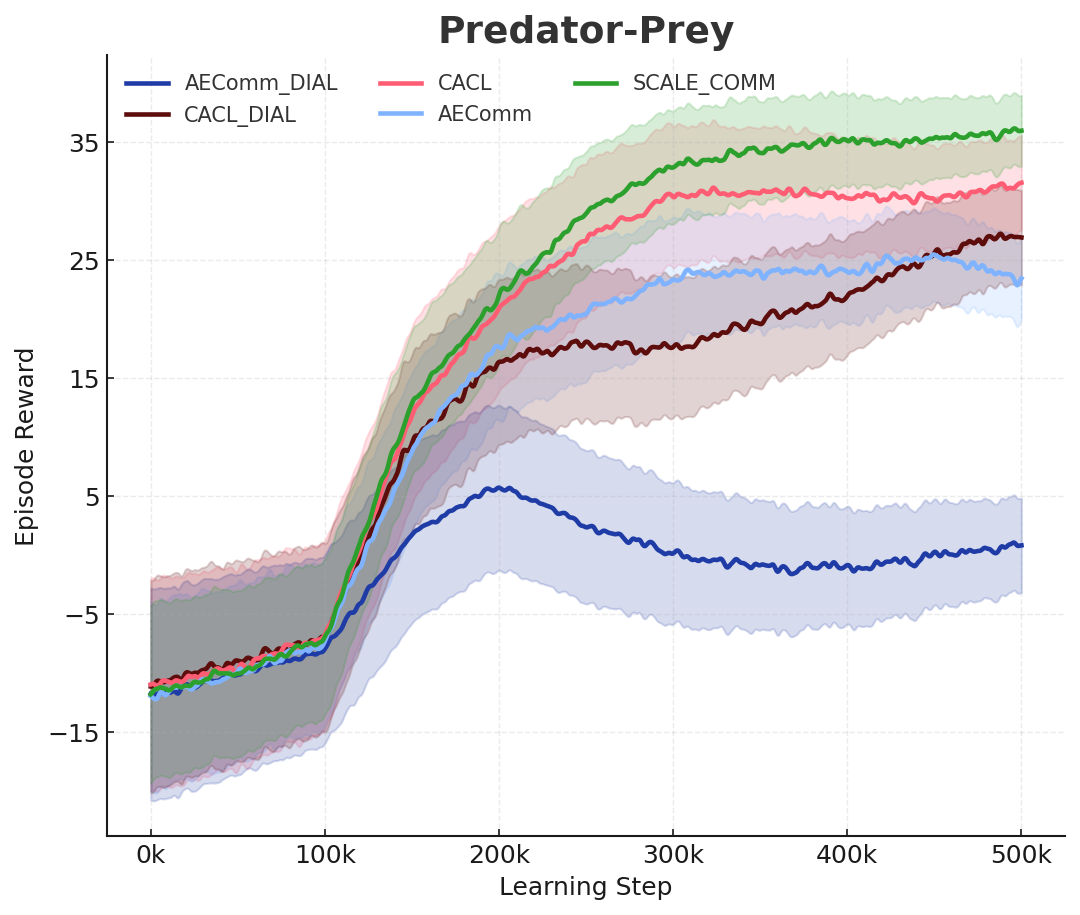}%
    \label{fig:predator}%
  }\hfill
  \subfloat[Find-Goal]{%
    \includegraphics[width=0.31\textwidth]{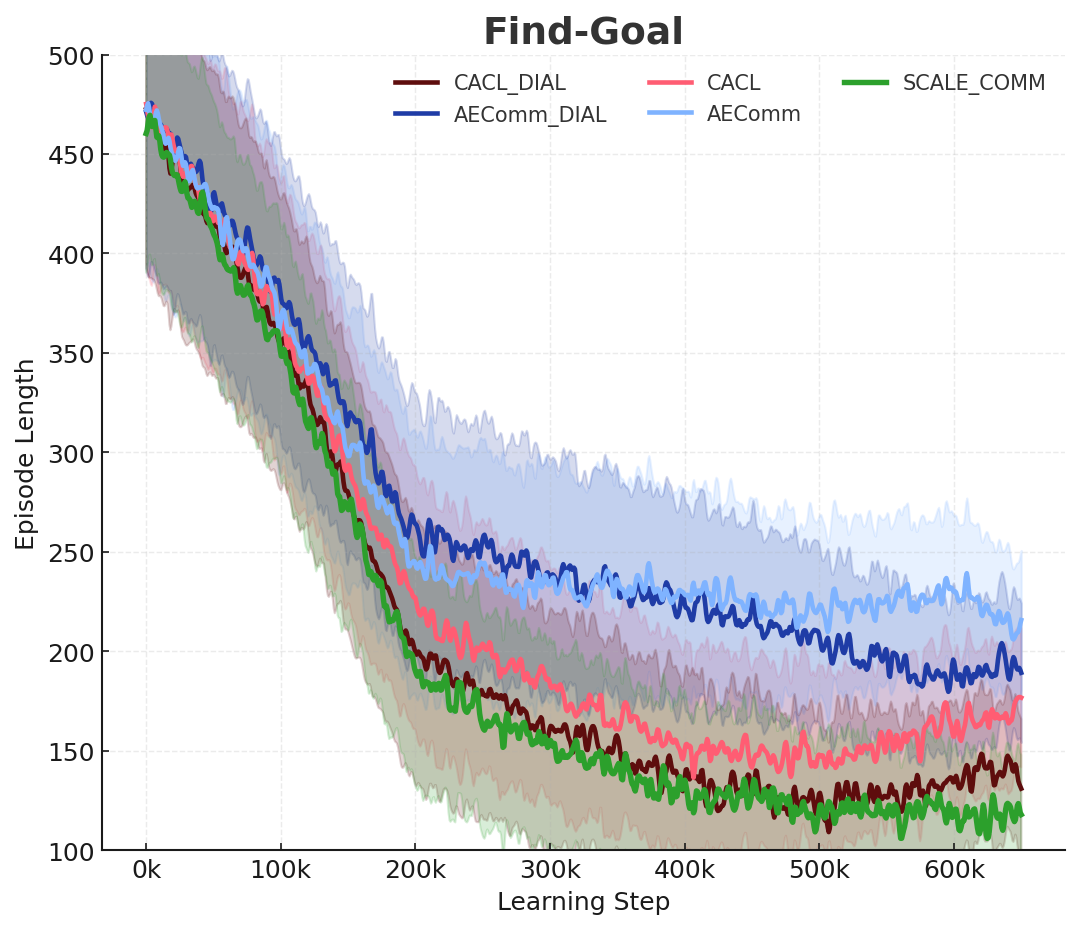}%
    \label{fig:findgoal}%
  }
  \caption{Performance comparison of \textbf{SCALE-COMM} and baseline methods across three cooperative multi-agent environments. (a)~Traffic-Junction: success rate (\%). (b)~Predator-Prey: episode reward (higher is better). (c)~Find-Goal: episode length (lower is better). Shaded regions denote 95\% confidence intervals across five random seeds.}
  \label{fig:benchmark_results}
\end{figure*}

We evaluated and compared SCALE-COMM against four baselines (AEComm, CACL, AEComm-DIAL, and CACL-DIAL) across three cooperative multi-agent environments: Traffic-Junction, Predator-Prey, and Find-Goal. Traffic-Junction tests collision avoidance among agents navigating intersecting lanes; Predator-Prey assesses pursuit–evasion coordination in dynamic, partially observable settings; and Find-Goal measures goal-seeking efficiency in spatially distributed tasks. As shown in Figure~\ref{fig:benchmark_results}, SCALE-COMM achieved faster convergence and higher asymptotic performance than all baselines. In Traffic-Junction, it reached near-perfect success (over 99\%) in under 100 learning steps. In Predator-Prey, it attained the highest average episode reward ($\approx35$) while maintaining stability under stochastic predator behavior. In Find-Goal, where shorter episode length reflects improved cooperation, SCALE-COMM converged to $\approx120$ steps, outperforming CACL ($\approx175$), CACL-DIAL ($\approx140$), AEComm ($\approx215$) and AEComm-DIAL ($\approx200$).

\subsection{Communication Representation}

\begin{figure}[t]
  \centering
  \includegraphics[width=\columnwidth]{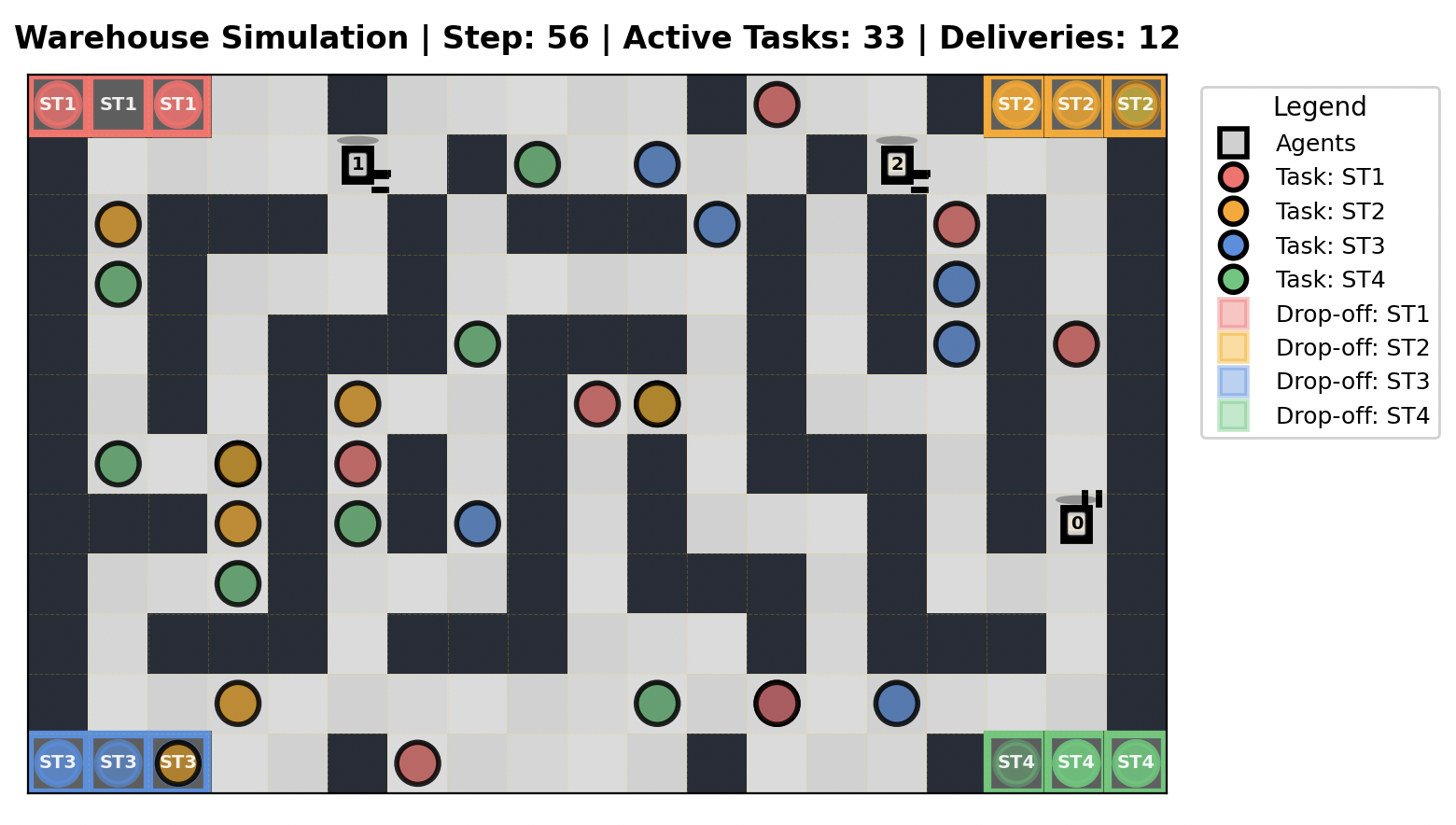}
  \caption{Example custom warehouse environment rollout. Agents (numbered squares) deliver tasks (colored circles) to drop-off corner locations to maximize throughput}
  \label{fig:warehouse_paths}
\end{figure}

We evaluated SCALE-COMM against SimCLR, MoCo, and SwAV in a warehouse task-allocation multi-agent environment (shown in Fig.~\ref{fig:warehouse_paths}) (40 episodes × 200 steps), training all models for 30 epochs with 192-dimensional embeddings under matched configurations. Representation quality was assessed using five complementary metrics: R@1 (cross-modal retrieval accuracy and preservation of semantic task identity), Temp@1 (short-horizon temporal consistency of adjacent agent states), ProtoNMI (coherence between self-organized prototype clusters and true task categories), ProbeAcc (downstream linear separability), and CKA(m,z) (structural similarity between intermediate activations and final embeddings, indicating architectural alignment and reduced redundancy). As shown in Table \ref{tab:results}, SCALE-COMM achieved near-perfect R@1, ProbeAcc, and CKA, and the highest ProtoNMI, with moderate Temp@1 improvements validating its temporal curriculum and neighbor-contrastive design. Ablation study is conducted against SCALE-COMM's three core representation components while auxiliary regularizers held fixed for stability: removing contrastive alignment primarily degrades instance retrieval (R@1) and structural alignment (CKA), while removing prototype distillation sharply reduces ProtoNMI and ProbeAcc, indicating loss of semantic structure. Removing curriculum scheduling causes smaller, uniform declines, suggesting it mainly stabilizes training rather than shaping representation geometry.

\begin{table}[t]
\centering
\scriptsize
\caption{Comparison of communication representation metrics across baselines (40 episodes, 30 epochs)}
\setlength{\tabcolsep}{4pt}
\renewcommand{\arraystretch}{1.05}
\begin{tabular}{@{}lccccccc@{}}
\hline
\textbf{Model} & \textbf{R@1} & \textbf{Temp@1} & \textbf{ProtoNMI} & \textbf{ProbeAcc} & \textbf{CKA(m,z)} \\ 
\hline
SimCLR       & 0.663 & 0.040 & 0.012 & 0.894 & 0.200 \\
\hline
MoCo         & 0.678 & 0.041 & 0.028 & 0.905 & 0.566 \\ 
\hline
SwAV         & 0.688 & 0.040 & 0.394 & 0.377 & 0.978 \\
\hline
\textbf{SCALE-COMM}  & \textbf{0.984} & \textbf{0.046} & \textbf{0.873} & \textbf{0.995} & \textbf{0.999} \\ 
\quad w/o Contrastive Alignment
         & 0.712 & 0.041 & 0.801 & 0.934 & 0.862 \\
\quad w/o Prototype Distillation
         & 0.961 & 0.045 & 0.392 & 0.884 & 0.983 \\
\quad w/o Curriculum Scheduling
         & 0.912 & 0.040 & 0.805 & 0.927 & 0.934 \\
\hline
\end{tabular}
\label{tab:results}
\end{table}

\subsection{Throughput and Efficiency Analysis}

The same custom environment was used to compare SimCLR, MoCo, SwAV, and SCALE-COMM representations (3 agents, 300-step episodes). All policies used identical PPO hyperparameters (6 epochs per iteration, 16,384 steps per iteration). As shown in Table \ref{tab:throughput}, SCALE-COMM achieved the highest throughput (37.20 deliveries/episode) and lowest unassigned time, indicating more efficient coordination. Qualitatively, the SCALE-COMM policy exhibits more structured and anticipatory routing behavior, where agents coordinate pickup and drop-off paths that minimize detours and congestion. This improvement stems from SCALE-COMM’s temporally consistent and communication-aware representation, which enhances task selection under uncertainty. The task-attention and proto-affinity modules bias actions toward spatially and temporally coherent assignments, leading to reduced "unassigned" movement and higher delivery density within each episode.

\begin{table}[h]
    \centering
    \caption{Comparison of PPO performance across SSL initialization methods.}
    \label{tab:throughput}
    \begin{tabular}{lcc}
    \toprule
    \textbf{Method} & \textbf{Deliveries/ep ($\pm$SD)} & \textbf{Unassigned (\%)} \\
    \midrule
    SimCLR & 35.30 $\pm$ 2.57 & 44.46 \\
    MoCo & 35.10 $\pm$ 2.81 & 43.58 \\
    SwAV & 35.20 $\pm$ 2.04 & 45.56 \\
    \textbf{SCALE-COMM} & \textbf{37.20 $\pm$ 1.80} & \textbf{41.95} \\
    \bottomrule
    \end{tabular}
\end{table}

\section{Conclusion}
This work introduced SCALE-COMM, a self-supervised communication framework that unifies cross-agent contrast, temporal prediction, prototype abstraction, and invariance alignment into a single representation-driven architecture for multi-agent reinforcement learning. Across common benchmarks (Traffic-Junction, Predator-Prey, and Find-Goal) and a complex warehouse coordination task, SCALE-COMM consistently outperformed strong contrastive and clustering baselines. Its temporally consistent latent space yielded more stable inter-agent communication, higher throughput under PPO fine-tuning, and improved sample efficiency. While the proposed method achieves state-of-the-art performance, runtime analysis revealed that SCALE-COMM on average utilizes 82\% more CPU resources and 77\% more runtime than the other self-supervised baselines. This additional computational overhead is due to the cross-agent contrastive masking and prototype-affinity updates introduced during training. Future work will focus on optimizing this cost through asynchronous or distributed update schemes, lightweight encoder variants, and hybrid quantization techniques to preserve communication fidelity while improving computational efficiency.

\bibliographystyle{IEEEtran}
\bibliography{references}


\end{document}